\documentclass[10pt,twocolumn,letterpaper]{article}
\usepackage[pdftex]{graphicx}
\usepackage{color}

\usepackage[pagenumbers]{cvpr} % To force page numbers, e.g. for an arXiv version

\definecolor{cvprblue}{rgb}{0.21,0.49,0.74}
\usepackage[pagebackref,breaklinks,colorlinks,allcolors=cvprblue]{hyperref}

\def\paperID{26} % *** Enter the Paper ID here
\def\confName{CVPR}
\def\confYear{2025}

\title{Speed-up of Vision Transformer Models by Attention-aware Token Filtering}

\author{
Takahiro Naruko\\
Hitachi, Ltd.\\
{\tt\small takahiro.naruko.yf@hitachi.com}
\and
Hiroaki Akutsu\\
Hitachi, Ltd.\\
{\tt\small hiroaki.akutsu.cs@hitachi.com}
}

\begin{document}
\maketitle
\begin{abstract}
    Vision Transformer (ViT) models have made breakthroughs in image embedding extraction, which provide state-of-the-art performance in tasks such as zero-shot image classification.
    However, the models suffer from a high computational burden.
    In this paper, we propose a novel speed-up method for ViT models called Attention-aware Token Filtering (ATF).
    ATF consists of two main ideas: a novel token filtering module and a filtering strategy.
    The token filtering module is introduced between a tokenizer and a transformer encoder of the ViT model, without modifying or fine-tuning of the transformer encoder.
    The module filters out tokens inputted to the encoder so that it keeps tokens in regions of specific object types dynamically and keeps tokens in regions that statically receive high attention in the transformer encoder.
    This filtering strategy maintains task accuracy while filtering out tokens inputted to the transformer encoder.
    Evaluation results on retrieval tasks show that ATF provides $2.8\times$ speed-up to a ViT model, SigLIP, while maintaining the retrieval recall rate.
\end{abstract}

\section{Introduction}
Vision Transformer (ViT) models \cite{ViT, SigLIP, DeiT} exhibit remarkable performance in vision tasks such as image classification and text-to-image retrieval.
Transformers leverage attention mechanism to draw global dependencies \cite{transformer} within an image.
Another characteristic of ViT models is that they follow the scaling law \cite{scaling}; larger models provide better task performance.
However, larger models increase the computational burden and processing time.
This raises a need to speed-up the ViT models.

ViT models are sometimes used to extract embeddings related to specific types of objects.
For example, in a text-to-image retrieval where document images are retrieved by texts written on the images, ViT models are used to extract embeddings related to the texts.
Another example is industrial anomaly detection, where images of industrial products are classified into anomaly classes.
In this case, ViT models are expected to extract embeddings related to the products.

In this paper, we propose a novel speed-up method for ViT models, Attention-aware Token Filtering (ATF).
ATF filters out the input image of the model to the regions necessary to extract the embeddings.
The contributions of this paper are as follows:
\begin{itemize}
    \item Empirical analysis reveals that in shallow layers, some regions statically receive high attention in images.
    \item A novel speed-up method for ViT models, Attention-aware Token Filtering (ATF), is proposed.
          ATF introduces a novel token filtering module between a tokenizer and a transformer encoder of the ViT models without modifying or fine-tuning of the tokenizer or the transformer encoder.
          The filtering module keeps tokens in regions which statically receive high attention, in addition to regions where the objects are detected dynamically.
    \item Evaluation results show that ATF speeds-up a ViT model, SigLIP \cite{SigLIP}, by $2.8\times$ while maintaining the recall rate of text-to-image and image-to-text retrieval.
\end{itemize}

\section{Related Work}
ViT models share the same basic architecture across their variants \cite{ViT, SigLIP, DeiT}.
They consist of two components: a tokenizer and a transformer encoder.
The tokenizer converts an image into a set of vectors called tokens.
First, the tokenizer splits an image into patches.
Then, it linearly projects each patch into a vector independently of other patches.
Lastly, tokens are generated by adding each vector to another vector specific to the spatial position in the original image.
This tokenization makes the transformer, which was originally proposed for natural language processing \cite{transformer}, applicable to vision tasks.
The transformer encoder, which employs the transformer architecture \cite{transformer}, takes the tokens as input and extracts the embeddings.

Although ViT models provide remarkable task performance, they suffer from high computational cost and long processing time.
A-ViT \cite{AViT} improved the throughput of a ViT model by adaptively halting tokens inside the transformer.
It employs simple halting modules to calculate a halting probability per token in each layer \cite{AViT}.
Once the cumulative probability of a token reaches a threshold, the token is discarded from the computation.
Although A-ViT provided throughput improvements, it requires fine-tuning of the ViT model to compute the halting probability.
Another drawback is that modification of the transformer encoder is necessary to add the halting modules to each layer.
In contrast, our ATF does not require fine-tuning or modification of the transformer encoder because the token filtering module is introduced between the tokenizer and the transformer encoder.
This feature potentially makes ATF easily applicable to multiple foundational ViT models.

\section{Proposed Method}
\subsection{Analysis}
\begin{figure}[t]
    \centering
    \includegraphics[width=1.0\linewidth]{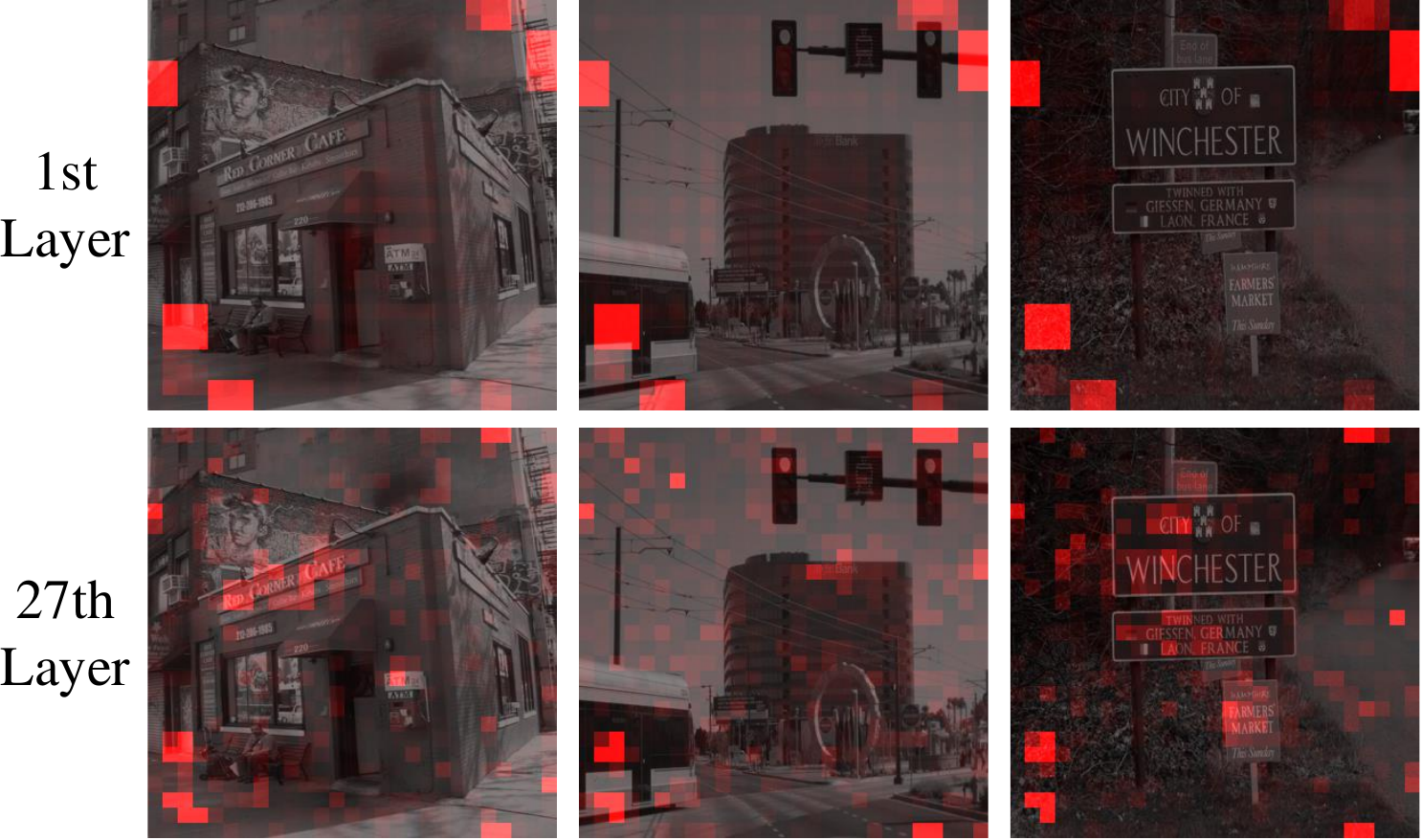}
    \caption{Regions with high attention values (red) on SigLIP \cite{SigLIP}.}
    \label{fig:attention_siglip}
\end{figure}
\begin{figure}[t]
    \centering
    \includegraphics[width=1.0\linewidth]{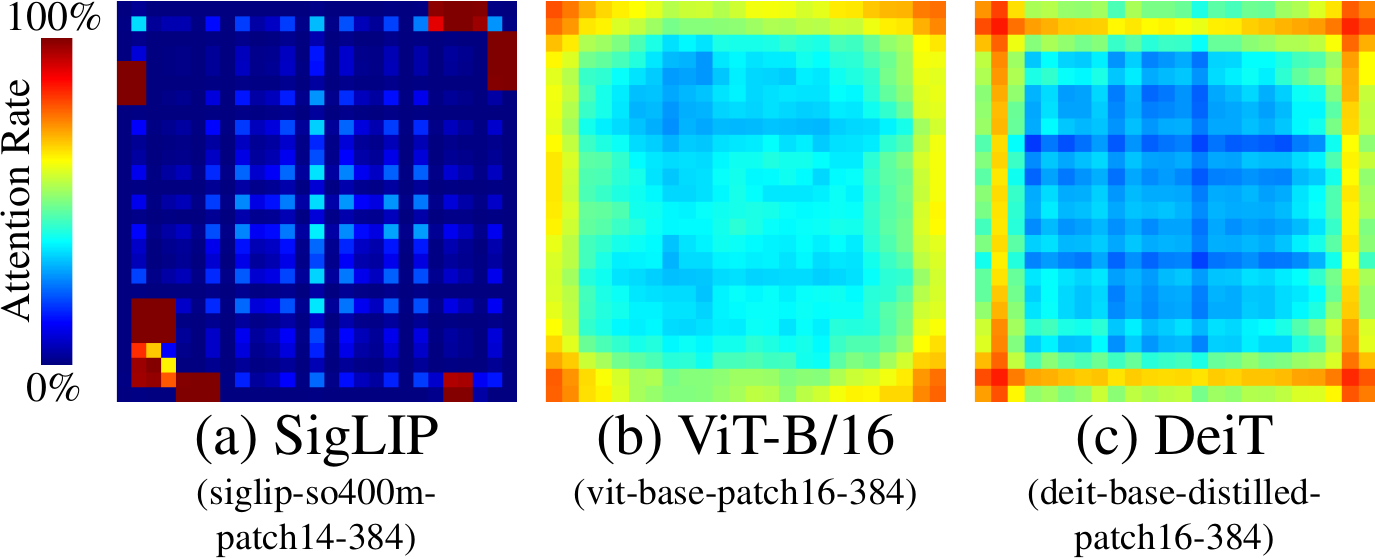}
    \caption{Visualization of attention rate in the 1st layer of SigLIP \cite{SigLIP}, ViT-B/16 \cite{ViT}, and DeiT \cite{DeiT}.}
    \label{fig:attention_rate}
\end{figure}
We empirically analyzed which regions in the image affect the embedding.
Assuming that tokens with higher attention values have larger impact on the embeddings if removed, we visualized attention values in each layer of the transformer encoder.
The experiments are conducted using images from the validation set of TextOCR dataset \cite{TextOCR}, the details of which are described later in Section \ref{sec:condition}, and the pretrained SigLIP \cite{SigLIP} model (siglip-so400m-patch14-384).
The results are shown in Figure \ref{fig:attention_siglip}, where the 1st and the 27th layers represent the shallowest and the deepest layers, respectively.
In the 1st layer, some specific regions receive high attention values in all images.
In contrast, as the layers go deeper to the 27th layer, regions with high attention values vary from image to image.

To discuss these findings more quantitatively, we define a metric called \textit{attention rate}, which represents how often each token receives high attention in the 1st layer.
We assume a token $t$ receives "high attention" in an image $i$ if its attention value $a_{i, t}$ is higher than the average value $\bar{a}_i$ in the image.
The attention value $a_{i, t}$ is calculated as follows:
\begin{equation}
    (a_{i, 1}, \cdots , a_{i, T}) = \frac{1}{H} \sum\nolimits_{h=1}^{H} \text{mean} (\text{softmax} (\frac{Q_{i, h} K_{i, h}^\top}{\sqrt{d_k}})),
\end{equation}
where $T$ is the number of tokens in an image, $H$ is the number of heads, $d_k$ is dimension of the keys, mean is a function to calculate average value along each column, and $Q_{i, h}$ and $K_{i, h}$ are query and key matrices of image $i$ in the $h$-th head, respectively.
Attention rate $r_t$ of a token $t$ is defined as $r_t = |\lbrace i \in I \mid a_{i, t} > \bar{a}_i\rbrace|/|I|$, where $I$ is a set of images from the validation set.

Figure \ref{fig:attention_rate} shows the visualization results of attention rate in the 1st layer of three ViT models: SigLIP \cite{SigLIP}, ViT-B/16 \cite{ViT}, and DeiT \cite{DeiT}.
Figure \ref{fig:attention_rate} (a) is the results on SigLIP \cite{SigLIP}.
In the figure, some areas are colored in dark red, which indicates that these regions receive high attention on almost all the images in the 1st layer of SigLIP.
This aligns with the findings obtained from Figure \ref{fig:attention_siglip}.
Figure \ref{fig:attention_rate} (b) and (c) show the results on ViT-B/16 \cite{ViT} and DeiT \cite{DeiT}, respectively.
These figures show that, although not contrastive as in SigLIP, ViT-B/16 and DeiT also have regions of high attention rate (e.g. corner areas in ViT-B/16 \cite{ViT} and edge areas in DeiT \cite{DeiT}).

In summary, the empirical analysis suggests that, in the shallow layers of ViT models, some specific regions statically receive high attention in images.

\subsection{Attention-aware Token Filtering (ATF)}
\begin{figure*}[t]
    \centering
    \includegraphics[width=1.0\linewidth]{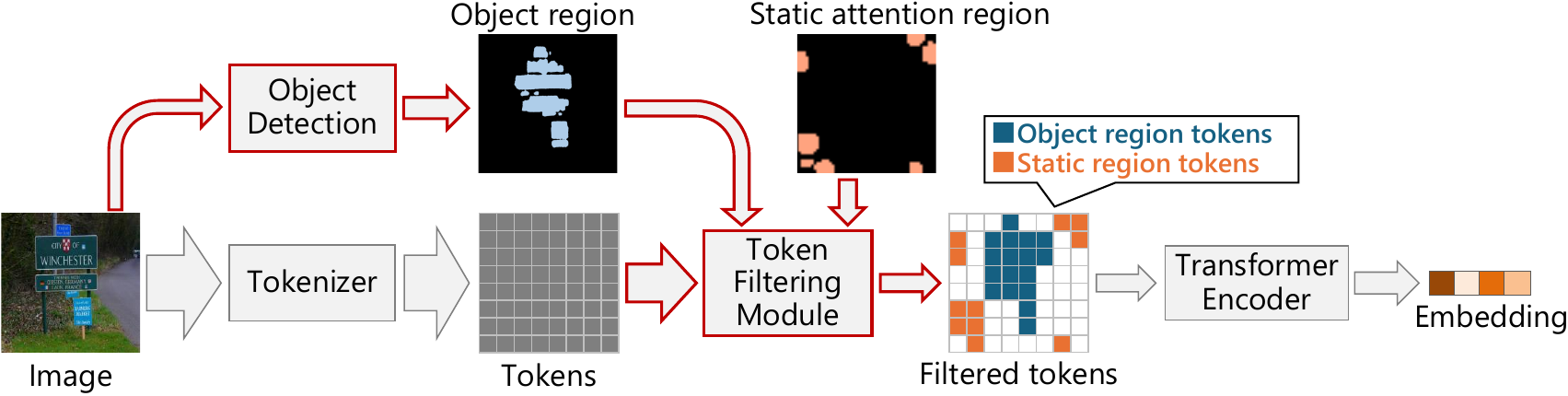}
    \caption{Architectural overview of proposed Attention-aware Token Filtering (ATF).}
    \label{fig:arch}
\end{figure*}
We propose a novel speed-up method for ViT models, Attention-aware Token Filtering (ATF).
ATF aims to accelerate ViT models used to extract embeddings related to specific object types.
An architectural overview of ATF is shown in Figure \ref{fig:arch}.
ATF consists of two main ideas.
\begin{enumerate}
    \item Token filtering module is newly introduced between the tokenizer and the transformer encoder.
    \item Novel token filtering strategy is used to filter out tokens without task accuracy drop.
\end{enumerate}

One approach to speed-up ViT models is reducing the amount of tokens processed in the transformer encoder, which is the bottleneck of the model.
There are some options when to reduce the tokens.
The previous work \cite{AViT} proposed a method adaptively discarding tokens inside the transformer encoder, which in turn requires modification and fine-tuning of the transformer encoder.
Another option is to crop an image so that the cropped one covers regions of the objects.
However, because the cropped image must be rectangular, there remain regions without the objects.
Our ATF inserts a token filtering module between the tokenizer and the transformer encoder.
This design choice frees both the tokenizer and the transformer encoder from modification and fine-tuning.
Moreover, it can filter out tokens without being constrained by structural assumptions as in 2D images because tokens are just a set of independent vectors converted from patches.

According to the empirical analysis mentioned above, the transformer encoder statically pays attention in the shallow layers to the same regions of tokens that we call \textit{static region tokens}.
Because it is assumed that removal of tokens with high attention values affects the extracted embeddings, our filtering strategy keeps these tokens.
More specifically, we define static region tokens as tokens that receive an attention value in the 1st layer that is higher than the value if all tokens receive the same attention equally.
%That is, $t$-th token is selected as the static region token if and only if the following condition holds in the 1st layer:
That is, a mask vector $\boldsymbol{s} = (s_1,\cdots ,s_T)$ to represent indices of static region tokens is defined as follows:
\begin{equation}
    s_t = 
        \begin{cases}
            1 & \text{if $\frac{1}{|I_s|} \sum\nolimits_{i \in I_s} a_{i, t} > \frac{1}{T|I_s|} \sum\nolimits_{t=1}^T \sum\nolimits_{i \in I_s} a_{i, t}$} \\
            0 & \text{otherwise}
        \end{cases}
    \label{eq:srt}
\end{equation}
where $I_s$ is a set of sample images.
%If the ViT model employs multi-head attention, $a_{i, t}$ is obtained by averaging the attention values over the heads.
Since indices of static region tokens are the same on all images by definition, they can be selected based on attention values in sample images $I_s$ prior to testing.
Thus, ATF does not need to choose static region tokens for each test image.

The analysis also suggested that the static region tokens are not enough to extract the embeddings because high attention tokens vary in the deep layers.
Since we aim to extract embeddings related to specific object types, it is assumed that tokens in regions of these objects are necessary.
Thus, ATF detects the object regions and keeps tokens of patches that contain the detected regions.
We call these tokens \textit{object region tokens}.
The object regions can potentially be detected in any way: with off-the-shelf semantic segmentation models \cite{PSPNet}, few-shot semantic segmentation models \cite{SSP}, and other dedicated models trained for the object types.

In summary, when ATF is applied to a ViT model with tokenizer $\mathcal{T}$ and transformer encoder $\mathcal{E}$, embedding $\boldsymbol{e}$ of an image $\boldsymbol{i}$ is extracted as follows:
\begin{equation}
    \boldsymbol{e} = \mathcal{E}(f(\mathcal{T}(\boldsymbol{i}), \boldsymbol{s} \lor D(\boldsymbol{i}))),
\end{equation}
where $f((\boldsymbol{x}_1,\cdots , \boldsymbol{x}_T), (m_1,\cdots , m_T))$ is the token filtering module that keeps token $\boldsymbol{x}_t$ if $m_t=1$, and $D$ is the object detection, which returns a mask vector $\boldsymbol{d}$ set to $1$ if the objects are detected in the token.

\section{Experiments}
\begin{table*}[t]
    \centering
    \caption{Evaluation results on text-to-image retrieval task using TextOCR dataset. SigLIP \cite{SigLIP} is used as a ViT model for both with and without ATF. Values in the parentheses are processing time of the object detection and the ViT model, respectively.}
    \label{table:t2i}
    \begin{tabular}{c|cccc|cccc|c|c}
        \toprule
        Method               & \multicolumn{4}{|c|}{Text-to-Image Retrieval Recall@$K$}  & \multicolumn{4}{|c|}{Image-to-Text Retrieval Recall@$K$}   & Processing time  & \# of \\
                             & $K=1$  & $5$    & $10$   & Avg.                           & $K=1$  & $5$    & $10$   & Avg.                            & [ms/image]       & tokens \\
        \midrule
        without ATF          & 87.6\% & 94.6\% & 95.7\% & 92.7\%                         & 86.1\% & 94.0\% & 95.9\% & 92.0\%                          & 268.0 (0.0 + 268.0)            & 2916 \\
        with ATF             & 87.5\% & 94.4\% & 95.9\% & 92.6\%                         & 86.0\% & 94.2\% & 95.9\% & 92.0\%                          & 94.4 (15.7 + 78.8) & 1190 \\
        \bottomrule
    \end{tabular}
\end{table*}
\begin{table}[t]
    \centering
    \caption{Ablation study of keeping static region tokens (SRT).}
    \label{table:srt}
    \begin{tabular}{c|cccc|c}
        \toprule
        SRT & \multicolumn{4}{|c|}{Text-to-Image Retrieval Recall@$K$}    & \# of  \\
            & $K=1$  & $K=5$  & $K=10$ & Average  & tokens \\
        \midrule
        No  & 54.9\% & 65.2\% & 68.7\% & 62.9\%   & 1012 \\
        YES & 87.5\% & 94.4\% & 95.9\% & 92.6\%   & 1190 \\
        \bottomrule
    \end{tabular}
\end{table}

\subsection{Experimental conditions}
\label{sec:condition}
We evaluated the effectiveness of our ATF with text-to-image and image-to-text retrieval task on TextOCR dataset \cite{TextOCR}.
The dataset consists of images containing texts, annotations representing the text regions, and ground truth texts written on the images.
3104 images from the validation set were used for the evaluation, omitting 18 images without ASCII texts or images with too large resolution.
Queries were generated using the following template:
\begin{center}
    \textit{An image with text "[text 1]", …, "[text 5]"},
\end{center}
where [text $i$] is a ground truth text that has the $i$-th largest area.
The task results were evaluated with top-1, 5, and 10 recall rate (denoted by Recall@$K$, where $K=1, 5, 10$).
As a ViT model, we used SigLIP \cite{SigLIP} (siglip-so400m-patch14-384), which originally accepts images with resolution $384\times384$.
Since the TextOCR \cite{TextOCR} dataset contains images with small texts, we fine-tuned SigLIP so that it can receive images with higher resolution of $768\times768$.
We trained only the vision model and kept the text model frozen.
Starting from the pretrained model, we trained the model with images and queries from the training set of TextOCR \cite{TextOCR} for 1 epoch.
Learning rate, $\beta_1$ of Adam optimizer, and weight decay were set to $1\times10^{-5}$, $0.85$, and $0.05$, respectively.
Note that the aim of this fine-tuning is to train the SigLIP model so that they can read higher resolution images, not to train the model towards our proposed ATF.
ATF itself does not require fine-tuning.
The same fine-tuned model was used to evaluate both SigLIP without ATF and SigLIP with ATF.

As sample images $I_s$ in Eq. \ref{eq:srt}, we used 128 images randomly sampled from the training set.
We trained an object detection model consisting of the first four layers of ResNet-50 \cite{ResNet} (i.e. from conv1 to conv4\_6), two transposed convolution layers, and interpolations.
Like in the fine-tuned SigLIP, test images were resized to $768\times768$ and inputted to the model at inference time.
The model was trained using images and annotations of the TextOCR \cite{TextOCR} training set.
Reducing false negatives is more crucial in ATF to maintain the retrieval recall rate.
Therefore, in the cross-entropy loss during training, we set 1 and 20 to the weights of the ground truth non-text and text areas, respectively.
At inference time, we set the softmax threshold to 0.4 and applied max pooling to expand the detected area by 12 pixels.

Processing times of the ViT model and the object detection model were measured on NVIDIA T4 GPU with FP16 precision.
Preprocessing such as image normalization and resize were not included.
Both models were executed using PyTorch 2.6.0 and Hugging Face transformers 4.49.0.

\subsection{Results}
The evaluation results of the retrieval tasks on TextOCR \cite{TextOCR} are summarized in Table \ref{table:t2i}.
"\# of tokens" shows the average number of tokens inputted to the transformer encoder per image.
The results demonstrate that SigLIP with ATF achieved a comparative recall rate as SigLIP without ATF, showing almost no performance degradation due to ATF.
%\textcolor{red}{That is, there is almost no performance degradation due to ATF.}
The results also show that ATF reduced the number of tokens from 2916 to 1190 ($1/2.5\times$) and processing time of the ViT model from 268 ms/image to 78.8 ms/image ($1/3.4\times$).
The reduction in the processing time ($1/3.4\times$) was greater than the reduction in the number of tokens ($1/2.5\times$) because transformers include attention mechanisms whose computational complexity is $O(T^2)$.
Although ATF needs object detection, its processing time was just 15.7 ms/image.
Therefore, the total processing time of SigLIP with ATF was 94.4 ms/image, which was $1/2.8\times$ smaller than that of SigLIP without ATF (268 ms/image).
In summary, ATF provided $2.8\times$ speed-up to the ViT model, SigLIP, while maintaining the retrieval recall rates.

To illustrate the effectiveness of the static region tokens, we evaluated the retrieval recall rate of ATF where static region tokens were disabled, i.e. only the object region tokens were inputted to the transformer encoder.
The ablation study results are summarized in Table \ref{table:srt}.
They show that without the static region tokens, the recall rate dropped by a large margin compared to the proposed ATF keeping these tokens.
They also show that the static region tokens increased the number of tokens by just 178, which was much less than 2916, the total number of tokens before filtering.
Therefore, ATF's token filtering strategy, which keeps the static region tokens, contributes to maintaining task accuracy with negligible token increase.

\section{Conclusion}
In this paper, we proposed a novel speed-up method for ViT models, Attention-aware Token Filtering (ATF).
ATF introduces a token filtering module between a tokenizer and a transformer encoder of the ViT models.
Therefore, ATF does not need modification or fine-tuning of the tokenizer and the transformer encoder.
By empirical analysis, we revealed that there are regions which statically receive high attention in images.
Based on this analysis, ATF employs a token filtering strategy that keeps tokens in these regions (static region tokens), in addition to tokens in regions where the objects are detected dynamically.

Evaluation results showed that ATF speeds-up SigLIP \cite{SigLIP} by $2.8\times$ while maintaining the recall rate of the text-to-image and image-to-text retrieval task.
The results also showed the effectiveness of keeping the static region tokens.

The empirical analysis indicated that different types of ViT models have different degrees of how statically the tokens receive high attention.
In this paper, static region tokens were selected as tokens with attention higher than the average.
However, more sophisticated selection algorithms may increase the effectiveness and generality of ATF.
Investigation of such algorithms remains for future work.

{
    \small
    \bibliographystyle{ieeenat_fullname}
    \bibliography{main}
}


\end{document}